\pdfoutput=1
\documentclass[10pt,twocolumn,letterpaper]{article}

\usepackage{cvpr}              

\usepackage{algorithm}  
\usepackage{algorithmic} 
\usepackage{mathtools}
\usepackage{amsthm}
\usepackage{multirow}
\usepackage{makecell}
\usepackage{graphicx}
\usepackage{booktabs}
\usepackage[accsupp]{axessibility} 

\usepackage[dvipsnames]{xcolor}

\definecolor{cvprblue}{rgb}{0.21,0.49,0.74}
\usepackage[pagebackref,breaklinks,colorlinks,citecolor=cvprblue]{hyperref}

\def\paperID{4729} 
\def\confName{CVPR}
\def\confYear{2024}

\title{CroSel: Cross Selection of Confident Pseudo Labels \\ 
for Partial-Label Learning}

\author{Shiyu Tian\textsuperscript{1}
\quad
Hongxin Wei\textsuperscript{2}
\quad
Yiqun Wang\textsuperscript{1}
\quad
Lei Feng\textsuperscript{3}\thanks{Corresponding author.}\\
\textsuperscript{1}Chongqing University\\
\textsuperscript{2}Southern University of Science and Technology\\
\textsuperscript{3}Singapore University of Technology and Design\\
{\tt\small tiansy7@foxmail.com, weihx@sustech.edu.cn, yiqun.wang@cqu.edu.cn, lfengqaq@gmail.com}
}

\begin{document}
\maketitle
\begin{abstract}
Partial-label learning (PLL) is an important weakly supervised learning problem, which allows each training example to have a candidate label set instead of a single ground-truth label. Identification-based methods have been widely explored to tackle label ambiguity issues in PLL, which regard the true label as a latent variable to be identified. However, identifying the true labels accurately and completely remains challenging, causing noise in pseudo labels during model training. In this paper, we propose a new method called CroSel, which leverages historical predictions from the model to identify true labels for most training examples. First, we introduce a cross selection strategy, which enables two deep models to select true labels of partially labeled data for each other. Besides, we propose a novel consistency regularization term called co-mix to avoid sample waste and tiny noise caused by false selection. In this way, CroSel can pick out the true labels of most examples with high precision. Extensive experiments demonstrate the superiority of CroSel, which consistently outperforms previous state-of-the-art methods on benchmark datasets. Additionally, our method achieves over 90\% accuracy and quantity for selecting true labels on CIFAR-type datasets under various settings.
\end{abstract}    
\section{Introduction}
\label{sec:intro}

The past few years have seen an increased interest in deep learning due to its outstanding performance in various application domains, including image processing~\cite{chen2021imageprocession}, automatic driving~\cite{xiong2019autodriving}, and medical diagnosis~\cite{park2018medical}. The success of deep learning heavily relies on a massive amount of fully labeled data. However, it is challenging to obtain a large-scale dataset with completely accurate annotations in the real world. To address this challenge, many researchers have explored a promising weakly supervised learning problem called partial-label learning (PLL)~\cite{cour2011learningPLL2,feng2020provably,xu2021instance,wen2021LWS,wang2022pico,wu2022CRDPLL,hong2023long,qiao2023fredis, xu2023progressive,qiao2023decompositional}, where each training example to have a set of candidate labels that includes the true label. This problem arises in many real-world tasks such as automatic image annotation~\cite{chen2017autoimageannotation} and facial age estimation~\cite{panis2015faceage}.

As PLL focuses on multi-class classification, there is only one ground-truth label for each training example, and other labels in the candidate label set are actually wrong (false positive) labels, which would have a negative impact on model training. Therefore, there exists the challenge of \emph{label ambiguity} in PLL. To address this challenge, the current mainstream solution is to disambiguate the candidate labels so as to figure out the true label for each training instance~\cite{jin2002learningmultiplelabels,yu2016pllidentimaximum,liu2012PLL-idnbase,feng2020provably,lv2020proden}.
However, most of the existing disambiguation methods normally leverage simple heuristics to iteratively update the labeling confidences or pseudo labels~\cite{feng2018leveraging,feng2020provably,lv2020proden,wang2022pico,wu2022CRDPLL}, which could not achieve convincing performance in identifying the true label during the training phase. Generally, if more true labels of training instances can be identified, we can train a better model. 
This serves as our primary motivation for identifying as many true labels as possible for training instances, ultimately resulting in the creation of a desired model.

In this paper, we propose a method called \textbf{CroSel} (\textbf{Cro}ss \textbf{Sel}ection of Confident Pseudo Labels), which leverages historical prediction from deep neural networks to accurately identify true labels for most training examples. Our selecting criteria are based on the assumption that if a model consistently predicts the same label for an input image with high confidence and low volatility, then that label has a high probability of being the true label for that example. Using the cross selection strategy, the true labels of the vast majority of training examples can be accurately identified, with only negligible noise. Moreover, in order to avoid sample waste and tiny noise resulting from the selection, we also propose a co-mix consistency regularization to generate trainable targets for all examples. This regulation term serves as an essential complement to our method, which can further enhance the scope and accuracy of our selection of ``true" labels. The algorithm details are shown in Section \ref{section:method}.

Our main contribution can be summarized as follows:
\begin{itemize}
\item We propose a cross selection strategy to select confident pseudo labels in the candidate label set based on historical prediction. Our approach demonstrates a notable combination of high precision and ratio in selecting "true" labels within candidate label sets.
\item We introduce a new consistency regularization term that can leverage MixUp~\cite{zhang2017mixup} to enhance the data and generate trainable targets for all examples, serving as an important supplement to our method.
\item We experimentally show that CroSel achieves state-of-the-art performance on common benchmark datasets. We also provide extensive ablation studies to examine the effect of the different components of CroSel.
\end{itemize}

\section{Related Work}
\label{sec:Related Work}

\begin{figure*}[!th]
\begin{center}
\centerline{\includegraphics[width=1.0\textwidth]{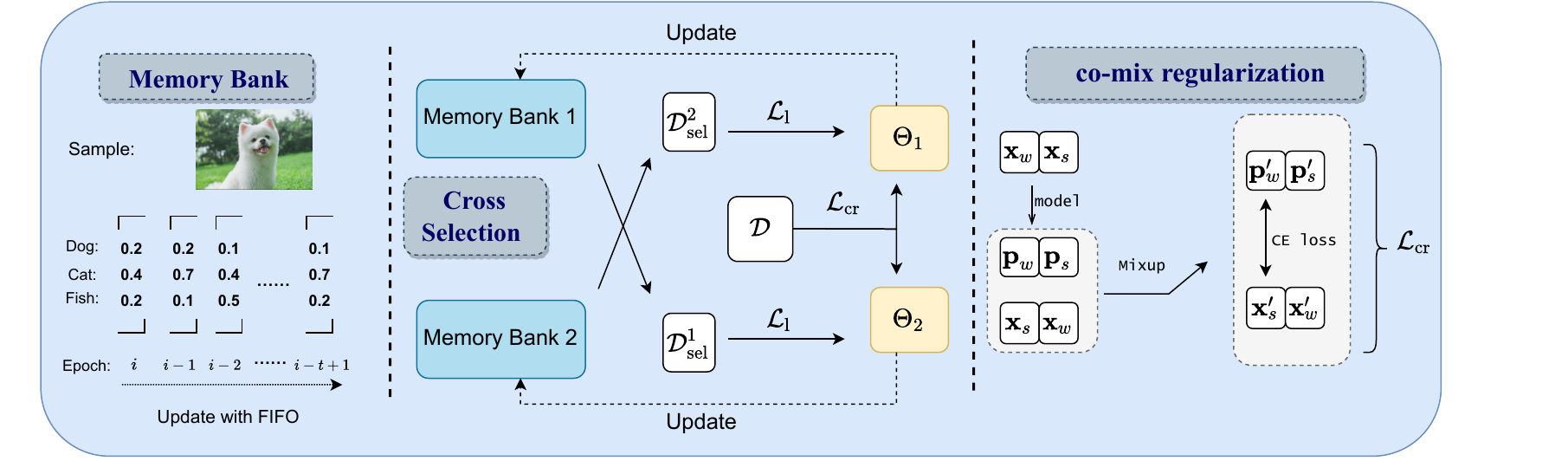}}
\caption{The left side of the figure is a brief example of our memory bank(MB) that stores the softmax output of the model for the last $t$ epochs, which is updated by the FIFO (First In First Out) principle; the middle is the cross selection strategy: within each epoch, data subsets $\mathcal{D}_{\mathrm{sel}}$ with confident pseudo-labels are selected from the MB of each network, which produce loss function $\mathcal{L}_{\mathrm{l}}$ to the training process for the other network;
the right side illustrates our co-mix regularization term and the corresponding loss function $\mathcal{L}_{\mathrm{cr}}$ .}
\label{fig:overview}
\end{center}
\vspace{-0.5cm}
\end{figure*}

\noindent \textbf{Partial-label learning.\quad} This setting allows each training example to be annotated with a set of candidate labels for which the ground truth label is guaranteed to be included. However, \emph{label ambiguity} can pose a significant challenge in PLL. Early methods use an averaging strategy, which tends to treat each candidate label equally.~\cite{cour2011learningPLL2,zhang2017plldisambiguation} But such methods are easily affected by negative labels in the candidate label set, thus forming wrong classification boundaries. Afterwards, identification-based methods ~\cite{jin2002learningmultiplelabels,yu2016pllidentimaximum,liu2012PLL-idnbase} have received more attention from the community, which regard the ground-truth label as a latent variable, and will maintain a confidence score for each candidate label. For instance, Yu \emph{et al.}~\cite{yu2016pllidentimaximum} introduced the maximum margin constraint to PLL, trying to optimize the margin between the model outputs from candidate labels and other negative labels. 


Recently, partial-label learning has been combined with deep networks, leading to significant improvements in performance. Feng \emph{et al.}~\cite{feng2020provably} assumed that partial labels come from a uniform generation model and provided a mathematical formulation, which is adopted by most of the algorithms proposed later~\cite{wang2022pico,xu2023pop,he2023candidateselect,wen2021LWS}. Based on this, they also propose an algorithm for classification consistency and risk consistency. PRODEN~\cite{lv2020proden} assumes the true label should be the one with the smallest loss among the candidate labels, and improves the classification risk algorithm accordingly. Wen \emph{et al.}~\cite{wen2021LWS} proposed a risk-consistent leveraged weighted loss with label-specific candidate label sampling.
PiCO~\cite{wang2022pico} innovatively introduces contrastive learning~\cite{oord2018contrastive} to the field and provides a solid theoretical analysis based on the EM (Expectation-Maximum) algorithm.
POP~\cite{xu2023pop} updates
the model and purifies each candidate label set progressively in every epoch and eventually approximates the Bayes optimal classifier with mild assumptions.

\noindent\textbf{Sample selection.\quad} Sample selection is a popular technique in deep learning, especially for learning with noisy labels \cite{liu2015classification,patrini2017making,zhang2018generalized,xia2019anchor,yao2020dual,wei2022smooth,wei2022learning,cheng2023mitigating}. Then an obvious idea is to separate the clean samples and noisy samples in the mixed dataset. 
To address this issue, many existing works adopt the small loss criterion~\cite{han2018coteaching,jiang2018mentornet,wei2020jocor}, which assumes that clean samples tend to have a smaller loss than noisy samples during training.
MentorNet~\cite{jiang2018mentornet} is a representative work that makes the teacher model pick up clean samples for the student model. Co-teaching~\cite{han2018coteaching} constructs a double branches network to select clean samples for each branch, which is different from the teacher-student approach since none of the models supervise the other but rather help each other out. This idea was improved by some research later~\cite{yu2019coteachingimprove1,wang2019coteachingimprove2}.
Curriculum learning is also applied to this field~\cite{han2018CLlabelnoise1}, which considers clean labeled data as an easy task, while noisily labeled data as a harder task. 
Guo \emph{et al.}~\cite{guo2018curriculumnet} splitted data into subgroups according to their complexities, in order to optimize the training objectives in the early stage of course learning. 
OpenMatch~\cite{saito2021openmatch} trains $n$ OVA classifiers to select the in-distribution samples under the open set setting. 
Generative models such as Beta Mixture Model ~\cite{arazo2019betamixture} and Gaussian Mixture Model~\cite{li2020dividemix} are also used to fit loss functions to distinguish clean labels from noisy labels. Recently, the fluctuation magnitude of the output~\cite{wei2022self,yuan2023late} and normalized entropy~\cite{he2023candidateselect} are also considered as an important credential to judge whether the label is clean.

\section{Our Proposed Method}\label{section:method}

In this section, we provide a detailed explanation of how our algorithm works. Our method is composed of two main components: a cross selection strategy that utilizes two models to select confident pseudo labels for each other, and a consistency regularization term that is applied across different data augmentation versions. The latter part not only addresses the issue of label waste resulting from the selection process but also enhances the quantity and accuracy of selections. The pseudo-code for our algorithm is presented in Algorithm \ref{alg:crosel}.

\subsection{Problem Setting}
Suppose the feature space is $\mathcal{X}\in\mathbb{R}^d$ with $d$ dimensions and the label space is $\mathcal{Y}=\{1,2,\dots,k\}$ with $k$ classes. We are given a dataset $\mathcal{D} =\{(\boldsymbol{x}_i,S_i)\}_{i=1}^n$ with n examples,
where the instance $\boldsymbol{x}_i\in\mathcal{X}$ and corresponding candidate label set $S_i\subset\mathcal{Y}$. Same as previous studies, we assume that the true label $y_i\in\mathcal{Y}$ of each input $\boldsymbol{x}_i$ is concealed in $S_i$.

Our aim is to train a multi-class classifier $f:\mathbb{R}^d\rightarrow\mathbb Y$ that minimizes the classification risk on the given dataset. For our classifier $f$, we use $f(\boldsymbol{x})$ to represent the output of classifier $f$ on given input $\boldsymbol{x}$. And we use $\hat{y}=\mathrm{argmax}_{y\in\mathcal{Y}}f_y(\boldsymbol{x})$ to denote the prediction of our classifier, where $f_y(\boldsymbol{x})$ is the $y\text{-}$th coordinate of $f(\boldsymbol{x})$.

\begin{algorithm*}[tb]
    \centering
   \caption{Pseudo-code of CroSel}
   \label{alg:crosel}
\begin{algorithmic}
   \STATE {\bfseries Input:} Training dataset $\mathcal{D}=\{(\boldsymbol{x}_i,S_i)\}_{i=1}^n$, consistency regularization parameter $\lambda_{\mathrm{cr}}$, sharpen parameter $T$, confidence threshold $\gamma$, memory bank $\mathrm{MB}^{(1)}$, $\mathrm{MB}^{(2)}$, network $\Theta^{(1)},\Theta^{(2)}$, epoch $E$, iteration $I$.
   \STATE {\bfseries Procedure:}
   \STATE $\mathrm{MB}^{(1)},\mathrm{MB}^{(2)},\Theta^{(1)},\Theta^{(2)}=\mathrm{WarmUp}(\mathrm{MB}^{(1)},\mathrm{MB}^{(2)},\Theta^{(1)},\Theta^{(2)},\mathcal{D})$. \qquad // \quad CC algorithm
   \FOR{$e=1$ {\bfseries to} $E$}
   \STATE Select labeled dataset $\mathcal{D}_{\mathrm{sel}}^1$ through $\mathrm{MB}^{(1)}$.     
   \STATE Select labeled dataset $\mathcal{D}_{\mathrm{sel}}^2$ through $\mathrm{MB}^{(2)}$.\qquad // \quad Eq. (4)
    \FOR{$k=1,2$ }
    \FOR{$i=1$ {\bfseries to} $I$}
    \STATE   Fetch a labeled batch $\hat{B}_i$ from the opposite selected dataset $\mathcal{D}_{\mathrm{sel}}^{\thicksim k}$.
    \STATE   Fetch a  batch $B_i$ from the training dataset $\mathcal{D}$.
    \STATE   Calculate the loss $L$ among the two batches $\hat{B}_i$ and $B_i$ through Eq. (13).
    \STATE   Update the weight of $\Theta^{(k)}$ by optimizer.
    \ENDFOR    
    \STATE  Update the memory bank $\mathrm{MB}^{(k)}$ through the FIFO principle.
    \ENDFOR
   \ENDFOR
\end{algorithmic}
\end{algorithm*}

\subsection{Selection Strategy}
In the current partial-label learning task, the large number of candidate labels can confuse the classifier, making it difficult for the classifier to capture specific features belonging to a certain label. Therefore, our goal is to identify the most likely true label among the candidate labels and eliminate the interference of other negative labels during model training. By selecting these ``true" labels, we can train the data with a supervised learning approach.

\noindent\textbf{Warm up.\quad}Before selecting, we warm up the network using the entire training set. The goal of this stage is to reduce the classification risk of the input $x$ to the whole candidate label set $S$, and obtain some historical information that can be used for selection. Therefore, here we use CC algorithm~\cite{feng2020provably} to warm up models for 10 epochs.
At the same time, we will update the value of memory bank $\mathrm{MB}$.

\noindent\textbf{Selecting criteria.\quad}We have three criteria for selecting the confident pseudo labels from the candidate label set.
Depending on the setup of our problem, the true label of each example must be in its candidate label set. This is our first criterion. In addition, we believe that an example's predicted label is likely to be the true label if the model predicts it with high confidence and low fluctuation. To determine the latter, we maintain a memory bank $\mathrm{MB}$ to store historical prediction of this neural network.

The size of the memory bank is $t\times n \times k$, where $t$ denotes the length of time it stores, $n$ is the length of dataset, and $k$ is the number of categories in the classification. In other words, $\mathrm{MB}$ stores the output after softmax of the model in the last $t$ epochs. 
$\mathrm{MB}$ is structured as a queue, with each element being a $k$-dimensional vector $\boldsymbol{q}$ representing the output of an example in a particular epoch. We update $\mathrm{MB}$ with the FIFO principle.
These selection criteria can be summarized as follows:
\begin{align}
\beta_1 &= \mathbb{I}(\mathrm{argmax}(\boldsymbol{q}^i) \in S),\\
\beta_2 &= \mathbb{I}(\mathrm{argmax}(\boldsymbol{q}^i) = \mathrm{argmax}(\boldsymbol{q}^{i+1})),\\
\beta_3 &= \mathbb{I}(\frac{1}{t} \sum\nolimits_{i=1}^{t} \mathrm{max}(\boldsymbol{q}^i) > \gamma),
\end{align}
where $\gamma$ is the confidence threshold of selection, $\mathbb{I}(\cdot)$ is the indicator function.
$\beta_1$ lets our selected label be in the candidate label set. $\beta_2$ limits that the label we picked has not flipped in the past $t$ epochs, which is a volatility consideration, and $\beta_3$ ensures that the label we selected has a high confidence level. The final selected dataset is $\mathcal{D}_{\mathrm{sel}}$:
\begin{equation}
\mathcal{D}_{\mathrm{sel}} = {((\boldsymbol{x}_i,\mathrm{argmax}(\boldsymbol{q}^t_i)) | (\beta_1^i \wedge \beta_2^i \wedge \beta_3^i) =1, \boldsymbol{x}_i \in \mathcal{D})},
\end{equation}

\noindent\textbf{Selected label loss.\quad}
After selecting the high-confidence examples, we obtain a dynamically updated data subset $\mathcal{D}_{\mathrm{sel}}=(X,\hat{Y})$. Each tuple of the subset has an instance $\boldsymbol{x}$ and a selected label $\hat{y}=\mathrm{argmax}(\boldsymbol{q}^t)$. In this configuration, we can use the basic cross-entropy loss to deal with this part of the samples:
\begin{equation}
\mathcal{L}_{\mathrm{l}} =\frac{1}{|\mathcal{D}_{\mathrm{sel}}|} { \sum\nolimits_{\boldsymbol{x} \in \mathcal{D}_{\mathrm{sel}}}} \mathcal{L}_{\mathrm{CE}}(f(\boldsymbol{x}_{\mathrm{w}}),\hat{y}),
\end{equation}
where $\mathcal{L}_{\mathrm{CE}}(\cdot,\cdot)$ denotes the softmax cross entropy loss, $\boldsymbol{x}_{\mathrm{w}}$ denotes the weak augmented version of $\boldsymbol{x}$.

\noindent\textbf{Cross selection.\quad}
However, In the process of selecting labels, it is difficult to guarantee that the selected labels are 100\% accurate.
In order to maximize the accuracy of selection, we propose a cross selection framework based on the idea of ensemble learning. Specifically, we train two identical models $\Theta^{(1)}$ and $\Theta^{(2)}$ through the same training and label selection process. By forming different decision boundaries, the two models can adaptively correct most of the errors even if there is noise in the selected confident pseudo labels. 

To maximize the accuracy of the selected labels, we employ a cross-supervised training process. This involves using the selected dataset $\mathcal{D}_{\mathrm{sel}}^1$ from $\mathrm{MB}^{(1)}$ to train $\Theta^{(2)}$, and vice versa, using the selected dataset $\mathcal{D}_{\mathrm{sel}}^2$ from $\mathrm{MB}^{(2)}$ to train $\Theta^{(1)}$. By doing so, the two models can learn from each other and further improve their ability to select true labels.
In the test process, we will average the output of the two models to get the final prediction:
\begin{equation}
f^{\prime}(\boldsymbol{x}) =\frac{1}{2} (f^1(\boldsymbol{x})+f^2(\boldsymbol{x})),
\end{equation}
where $f^{\prime}(\boldsymbol{x})$ denotes the final output of our method in the test process, $f^{1}(\boldsymbol{x})$ denotes the output of $\Theta^{(1)}$, $f^{2}(\boldsymbol{x})$ denotes the output of $\Theta^{(2)}$.

\subsection{Co-mix Consistency Regulation}
\noindent\textbf{Motivation.\quad} When dealing with complex partial-label learning tasks, our label selection strategy may not accurately select the true labels for all examples. If we only use the selected examples with their corresponding labels, it will result in a significant amount of wasted data, which contradicts our goal of utilizing as many examples as possible. Therefore, we aim to provide a trainable target for the remaining examples that are not selected. 
However, our setting differs from traditional semi-supervised learning as the proportion of unlabeled examples is relatively small. So it is unreasonable to directly transfer the existing semi-supervised learning tools to unlabeled data like other weakly supervised learning methods. 

Motivated by this, we hope to propose a regularization term that can serve as an important supplement to our method and help us select examples.
We proposed the co-mix regularization term, which employs two widely used data augmentation methods: weak augmentation and strong augmentation, to generate pseudo-labels as training targets for consistency regularization. We further employ MixUp~\cite{zhang2017mixup} to further enhance the data. It is worth mentioning that the term ``pseudo labels" in this part specifically refers to the soft labels generated from different data augmentation versions, rather than the confident hard labels selected during the selection process.


\noindent\textbf{Pseudo label generating.\quad}Consistency loss is a simple but effective idea in weakly supervised learning, whose key point is to reduce the gap between the output of two perturbed examples after passing through the model. As discussed in the previous section, we used two widely used data augmentations: `weak' and `strong', and crossed them to generate pseudo labels. Specifically, as stated in Figure \ref{fig:overview}, the pseudo label corresponding to weak augmented example is generated by strong augmented example, while the pseudo label corresponding to strong augmented example is generated by weak augmented example.

To generate these pseudo labels, we fix the parameters of the neural network, and pass the augmented images through the model to get the logits output. Then we perform two operations on the logits, sharpening and normalization. For sharpening operation, we use a hyper-parameter $T$, the more $T$ goes to zero, the more logits tend to become a one-hot distribution. These two operations can be summarized by the following formula.
\begin{equation}
\boldsymbol{p}_{i}=
\begin{cases}
\frac{\exp(f_i(\boldsymbol{x})^{\frac{1}{T} })}{ {\textstyle \sum_{i\in S} \exp(f_i(\boldsymbol{x})^{\frac{1}{T} })} },& \text{$i \in S$},\\
0,& \text{$i \notin S$},
\end{cases}
\end{equation}
where $\boldsymbol{p}_{i}$ denotes the $i\text{-}\text{th}$ coordinate of pseudo label.

\noindent\textbf{MixUp.\quad} After generating the pseudo labels of each example, we end up with two datasets that can be trained: $(X_{\mathrm{w}},P_{\mathrm{s}})$ and $(X_{\mathrm{s}},P_{\mathrm{w}})$. The subscripts \emph{w} and \emph{s} indicate the type of data augmentation used, i.e., weak or strong. Then, We spliced the two datasets together to further enhance the data with MixUp~\cite{zhang2017mixup}. For a pair of two examples with their corresponding pseudo labels $(\boldsymbol{x}_1, \boldsymbol{p}_1),(\boldsymbol{x}_2, \boldsymbol{p}_2)$, we compute $(\boldsymbol{x}^{\prime}, \boldsymbol{p}^{\prime})$ by the following formula: 
\begin{align}
\lambda & \sim{\mathrm{Beta}(\alpha,\alpha)},
\\
\lambda^{\prime} &= \mathrm{max}(\lambda,1-\lambda),
\\
\boldsymbol{x}^{\prime}  &=\lambda^{\prime}\boldsymbol{x}_1+(1-\lambda^{\prime})\boldsymbol{x}_2,
\\
\boldsymbol{p}^{\prime}  &=\lambda^{\prime}\boldsymbol{p}_1+(1-\lambda^{\prime})\boldsymbol{p}_2,
\end{align}
After MixUp, we finally obtain $2n$ trainable example pair $(\boldsymbol{x}^{\prime}, \boldsymbol{p}^{\prime})$. Then we can use the typical cross-entropy loss on every example, the consistency regulation loss will be:
\begin{equation}
\mathcal{L}_{\mathrm{cr}} = \frac{1}{2n} { \sum\nolimits_{i=1}^{2n}} \mathcal{L}_{\mathrm{CE}}(f(\boldsymbol{x}^{\prime}_i),{\boldsymbol{p}^{\prime}_i}),
\end{equation}
where $\mathcal{L}_{\mathrm{CE}}(\cdot,\cdot)$ denotes the softmax cross entropy loss, $n$ denotes the length of a dataset.

\begin{table*}[t]
\centering
\caption{Accuracy (mean$\pm$std) comparisons on benchmark datasets.}
\label{Main reults}
\resizebox{1.00\textwidth}{!}{
\setlength{\tabcolsep}{3mm}{
\begin{tabular}{l|c|ccccccc}
\toprule
Dataset & $q$ & Ours & PoP &CRDPLL & PiCO & PRODEN & LWS & CC  \\
\midrule
\multicolumn{1}{c|}{\multirow{3}{*}{CIFAR-10}}    & $0.1$& 97.31$\pm$.04\% & 97.17$\pm$.01\% &\textbf{97.41$\pm$.06}\% & 96.10$\pm$.06\% &95.66$\pm$.08\% &91.20$\pm$.07\% &  90.73$\pm$.10\% \\
\multicolumn{1}{c|}{} & $0.3$& \textbf{97.50$\pm$.05}\% & 97.08$\pm$.01\% &97.38$\pm$.04\%  & 95.74$\pm$.10\% & 95.21$\pm$.07\% & 89.20$\pm$.09\% & 88.04$\pm$.06\% \\
\multicolumn{1}{c|}{}    & $0.5$& \textbf{97.34$\pm$.05}\% &96.66$\pm$.03\% &96.76$\pm$.05\% & 95.32$\pm$.12\% & 94.55$\pm$.13\% & 80.23$\pm$.21\% & 81.01$\pm$.38\% \\
\midrule
\multicolumn{1}{c|}{\multirow{3}{*}{SVHN}}    & $0.1$& \textbf{97.71$\pm$.05}\% & 97.55$\pm$.06\% &97.63$\pm$.06\% & 96.58$\pm$.04\% &96.20$\pm$.07\% &96.42$\pm$.09\% &  96.99$\pm$.17\% \\
\multicolumn{1}{c|}{} & $0.3$& \textbf{97.96$\pm$.05}\% & 97.50$\pm$.03\% &97.65$\pm$.07\%  & 96.32$\pm$.09\% & 96.11$\pm$.05\% & 96.15$\pm$.08\% & 96.67$\pm$.20\% \\
\multicolumn{1}{c|}{}    & $0.5$& \textbf{97.86$\pm$.06}\% & 97.31$\pm$.01\% &97.70$\pm$.05\% & 95.78$\pm$.05\% & 95.97$\pm$.03\% & 95.79$\pm$.05\%  & 95.83$\pm$.23\% \\
\midrule
\multicolumn{1}{c|}{\multirow{3}{*}{CIFAR-100}}    & $0.01$& \textbf{84.24$\pm$.09}\% & 83.03$\pm$.04\%&82.95$\pm$.10\% & 74.89$\pm$.11\% &72.24$\pm$.12\% &62.03$\pm$.21\% &  66.91$\pm$.24\% \\
\multicolumn{1}{c|}{} & $0.05$& \textbf{83.92$\pm$.24}\% & 82.79$\pm$.02\%&82.38$\pm$.09\%  & 73.26$\pm$.09\% & 70.03$\pm$.18\% & 57.10$\pm$.17\% & 64.51$\pm$.37\% \\
\multicolumn{1}{c|}{}    & $0.10$& \textbf{84.07$\pm$.16}\% & 82.39$\pm$.04\%&82.15$\pm$.20\% & 70.03$\pm$.10\% & 69.82$\pm$.11\% & 52.60$\pm$.54\%  & 61.50$\pm$.36\%\\
\bottomrule
\end{tabular}
}
}
\end{table*}

\subsection{Algorithm overview}

\noindent\textbf{Overall loss.\quad} In the formal training phase, our loss function will be composed of two parts, the supervised loss $\mathcal{L}_{\mathrm{l}}$ in the selected label set $\mathcal{D}_{\mathrm{sel}} $ and the consistency regularization item loss $\mathcal{L}_{\mathrm{cr}}$. The two will be dynamically combined into the final loss function by a hyperparameter:
\begin{equation}
\mathcal{L}_{\mathrm{all}} =  \mathcal{L}_{\mathrm{l}} + \lambda_{\mathrm{d}}*\mathcal{L}_{\mathrm{cr}},
\end{equation}
where $\lambda_{\mathrm{d}}$ is a dynamically changing parameter, $\mathcal{L}_{\mathrm{l}}$ and $\mathcal{L}_{\mathrm{cr}}$ can be calculated by Eq. (5) and Eq. (12), respectively. The use of MixUp can cause significant changes to the original feature space, making it necessary to adjust the weight of the regularization term in the loss function as the number of selected samples increases. 
To achieve this, we proposed a gradually decreasing $\lambda_{\mathrm{d}}$ with the increase of selected samples. We can control the magnitude of $\lambda_{\mathrm{d}}$ with a set hyperparameter $\lambda_{\mathrm{cr}}$. The updated rules of $\lambda_{\mathrm{d}}$ are as follows:
\begin{equation}
\lambda_{\mathrm{d}} = (1- r_{\mathrm{s}})* \lambda_{\mathrm{cr}},
\end{equation}
where $r_{\mathrm{s}}$ denotes the percentage of labeled data that we picked out, $\lambda_{\mathrm{cr}}$ is a hyperparameter that collaboratively adjusts the ratio of two loss items.
\section{Experiments}
\subsection{Experimental Setup}
\noindent\textbf{Datasets.\quad}We used three widely used benchmarks in this field: SVHN~\cite{svhn}, CIFAR-10~\cite{krizhevsky2009cifar} and CIFAR-100~\cite{krizhevsky2009cifar}.  The way we generate partial labels is by flipping the negative labels $\overline{y} \neq y$ of the example with a set probability $q=P(\overline{y} \in S| \overline{y} \neq y)$. With the increase of $q$, the noise of the dataset increases gradually. Following PiCO~\cite{wang2022pico}, we consider $q=\{0.01,0.05,0.1\}$ for CIFAR-100 and $q=\{0.1,0.3,0.5\}$ for other datasets.

\noindent\textbf{Compared methods.\quad} We choose six well-performed partial-label learning algorithms to compare: 
\begin{itemize}
    \item PoP~\cite{xu2023pop}, an algorithm that updates the learning model and purifies each candidate label set progressively in every epoch.
    \item CRDPLL~\cite{wu2022CRDPLL}, an algorithm that takes non-candidate labels as supervision information and proposes a new consistency loss term between augmented images.
    \item PiCO~\cite{wang2022pico}, a theoretical solid framework that combines contrastive learning and prototype-based label disambiguation algorithm.
    \item LWS~\cite{wen2021LWS}, an algorithm that wants to balance the risk error between the candidate label set and the non-candidate label set.
    \item PRODEN~\cite{lv2020proden}, a self-training algorithm that dynamically updates the confidence of candidate labels.
    \item CC~\cite{feng2020provably}, an algorithm that wants to minimize the classification error of the whole candidate label sets.
\end{itemize}

\begin{table}[!t]
\caption{Selection ratio and selection accuracy (mean$\pm$std) on benchmark datasets. S-ratio represents the selection ratio and S-acc represents selection accuracy in $\mathcal{D}_{\mathrm{sel}}$.}
\label{main scc+sration table}
\label{Co-selct results:Sratio and Sacc}
\resizebox{0.475\textwidth}{!}{
\setlength{\tabcolsep}{4mm}{
\begin{tabular}{c|c|c|c}
\toprule
\multicolumn{1}{l|}{Datasets} & Setting                  & Index          &  Performance\\ 
\midrule
\multirow{6}{*}{CIFAR-10}     & \multirow{2}{*}{$q=0.1$} & S-ratio &  99.09$\pm$.07\% \\
                              &                          & S-acc   &  99.79$\pm$.05\%\\ \cmidrule{2-4}
                              & \multirow{2}{*}{$q=0.3$} & S-ratio  &  98.10$\pm$.10\%\\
                              &                          & S-acc   &  99.55$\pm$.03\%\\ \cline{2-4}
                              & \multirow{2}{*}{$q=0.5$} &  S-ratio&  96.25$\pm$.12\%\\
                              &                          & S-acc  &  99.44$\pm$.06\%\\ \midrule
\multirow{6}{*}{SVHN}     & \multirow{2}{*}{$q=0.1$} & S-ratio &  97.25$\pm$.14\% \\
                              &                          & S-acc   &  99.84$\pm$.06\% \\ \cmidrule{2-4}
                              & \multirow{2}{*}{$q=0.3$} & S-ratio  &  76.42$\pm$.21\%\\
                              &                          & S-acc   &  99.77$\pm$.06\%\\ \cmidrule{2-4}
                              & \multirow{2}{*}{$q=0.5$} &  S-ratio&  73.21$\pm$.15\%\\
                              &                          & S-acc  &  99.34$\pm$.02\%\\ \midrule
\multirow{6}{*}{CIFAR-100}     & \multirow{2}{*}{$q=0.01$} & S-ratio &  96.58$\pm$.13\% \\
                              &                          & S-acc   &  99.71$\pm$.06\% \\ \cmidrule{2-4}
                              & \multirow{2}{*}{$q=0.05$} & S-ratio  &  95.45$\pm$.21\%\\
                              &                          & S-acc   &  98.29$\pm$.15\%\\ \cmidrule{2-4}
                              & \multirow{2}{*}{$q=0.10$} &  S-ratio&  93.61$\pm$.12\%\\
                              &                          & S-acc  &  97.93$\pm$.11\%\\ \midrule

\end{tabular}
}
}
\vspace{-0.4cm}
\end{table}

\noindent\textbf{Implementations.\quad} Our implementation is based on PyTorch~\cite{paszke2019pytorch}. We use WRN-34-10 (short for Wide-ResNet-34-10) as the backbone model with a weight decay of 0.0001 on all the datasets for all compared methods. We present the mean and standard deviation in each case based on three independent runs with different random seeds. More detailed hyper-parameter setting can be found in Supplementary Materials.


\subsection{Main Empirical Results}
As shown in Table \ref{Main reults}, our methods achieve state-of-the-art results on all the settings except CIFAR-10 with $q=0.1$. Notably, on the complex dataset CIFAR-100, our method significantly improves performance.

However, a counter-intuitive phenomenon appears in our experiment, that is, the performance of our method does not necessarily decline strictly with the increase of noise magnitude $q$; on the contrary, it may perform best in the case of moderate noise. We believe the possible reason for this phenomenon is that: On the one hand, when generating pseudo labels, we normalize them according to Eq. (7). That is to say, as the number of candidate labels increases, the ingredients involved in MixUp will also increase, leading to better-enhanced data interpolation. On the other hand, the increase of candidate labels also represents the increase of noise, which can have a negative impact on the algorithm's performance. Therefore, CroSel performs well in cases of intermediate noise, representing an optimal solution found in such a trade-off problem.

Table \ref{main scc+sration table} presents the selection ratio and accuracy of CroSel in selecting the true labels for the subdataset $\mathcal{D}_{\mathrm{sel}}$. It is evident that, CroSel can accurately select the true labels of most examples in the dataset under various noise conditions. Notably, for CIFAR-10 and CIFAR-100, the selection ratio and accuracy are over $90\%$ both. However, in SVHN, the selection ratio drops significantly with the increase of noise. This could be attributed to the fact that digital images in SVHN have relatively simple shape features, and MixUp may significantly disturb the feature space.

\noindent\textbf{More challenging experimental settings.\quad} We conducted fine-grained settings on CIFAR-100 by limiting the candidate labels to the super-class of the true label, which is closer to reality. We set $q = 0.5$, indicating that half of the remaining superclass labels for each example may be converted into candidate labels. Our method still exhibits superior performance in this scenario. The results can be found in Table \ref{Results on CIFAR-100 for fine-grained settings}. 

\begin{table}[!t]
\centering
\caption{Results on CIFAR-100 in fine-grained settings.}
\label{Results on CIFAR-100 for fine-grained settings}
\resizebox{0.475\textwidth}{!}{
\setlength{\tabcolsep}{5mm}{
\begin{tabular}{c|c|c|c}
\toprule
Method & Accuracy &Method & Accuracy\\
\midrule
PoP & 82.04\% & CRDPLL & 81.53\% \\
PiCO & 73.38\% & PRODEN & 71.16\% \\
LWS & 54.08\% & CC & 64.91\% \\
\midrule
\multicolumn{2}{c|}{Crosel (ours) } & \multicolumn{2}{c}{ \textbf{83.34\%} } \\
\bottomrule
\end{tabular}
}
}
\vspace{-0.3cm}
\end{table}

\subsection{Ablation Studies}
In this section, our primary objective is to demonstrate the collective effectiveness of all components within our method. Subsequently, we will delve into illustrating the consequences of either the absence or adjustment of certain components on the results.

\noindent\textbf{All the components matter.} Our algorithm comprises several components, including a selection strategy with three selection criteria and a regularization term. 
We conducted comprehensive ablation studies on CIFAR-100 with $q=0.1$ to further prove that every component of our algorithm matters. The results are shown in Table \ref{table:main ablation}. Notably, the absence of any component leads to a significant performance decline.
The Co-mix regularization term serves as a training target for the examples that are not selected which can improve the selection accuracy, selection number, and final test accuracy when utilizing the same selection criteria. 

Compared with using a combination of three criteria, employing individual selection criteria implies a relaxation of selection standards. When regularization terms are employed conventionally, this relaxation inevitably leads to decreased accuracy. While it enables the selection of more samples, it also introduces additional noise that can significantly misguide the model's training process.
In scenarios where no regularization terms are utilized, the imposition of more stringent selection criteria poses challenges in acquiring sufficient data for training. In such cases, relaxing the selection criteria serves to expand the model's training data, thereby enhancing the overall performance of the model and subsequently improving its test performance.




\begin{table}[!t]
\centering
\caption{Results of thorough ablation experiments.}
\vspace{-0.3cm}
\label{table:main ablation}
\resizebox{0.475\textwidth}{!}{
\setlength{\tabcolsep}{2.5mm}{
\begin{tabular}{cccc|ccc}
\toprule
cr1 & cr2 & cr3 & $\mathcal{L}_{\mathrm{cr}}$ & Acc & S-acc& S-ratio\\
\midrule
\checkmark & & & & 73.12\% & 95.15\% & 90.32\% \\
\checkmark & \checkmark& & & 72.00\% & 93.51\% & 85.09\% \\
\checkmark & \checkmark& \checkmark& & 70.68\% & 96.22\% & 78.65\% \\
\checkmark & & & \checkmark& 77.04\% & 91.02\% & 97.27\% \\
\checkmark & \checkmark& & \checkmark & 79.90\% & 94.66\% & 95.51\% \\
\checkmark &\checkmark &\checkmark & \checkmark& \textbf{84.07\%} & \textbf{97.93\%} & \textbf{93.61\%} \\
\bottomrule
\end{tabular}
}
}
\end{table}

\begin{table}[!t]
\centering
\caption{Results for ablation studies on the scope of regularization.}
\vspace{-0.3cm}
\label{Ablation scope}
\resizebox{0.475\textwidth}{!}{
\setlength{\tabcolsep}{3mm}{
\begin{tabular}{c|c|c|c}
\toprule
\multicolumn{1}{l|}{Setting} & Scope         & Index          &  Performance\\ 
\midrule
\multirow{9}{*}{\makecell[c]{CIFAR-10 \\ $q=0.5$}}     & \multirow{3}{*}{All data} & Acc &  \textbf{97.34\%} \\
                              &                          & S-ratio   &  \textbf{96.25\%}\\ 
                              &                          & S-acc   &  \textbf{99.44\%}\\ \cline{2-4}
                              & \multirow{3}{*}{Unselected data} & Acc  &  90.32\%\\
                              &                          & S-ratio   &  93.27\%\\ 
                              &                          & S-acc   &  95.72\%\\ \cline{2-4}
                              & \multirow{3}{*}{None} &  Acc&  81.01\%\\
                              &                          & S-ratio   &  90.23\%\\ 
                              &                          & S-acc  &  89.72\%\\ \hline
\multirow{9}{*}{\makecell[c]{CIFAR-100 \\ $q=0.1$}}     & \multirow{3}{*}{All data} & Acc &  \textbf{84.07\%} \\
                              &                          & S-ratio   &  \textbf{93.61\%}\\ 
                              &                          & S-acc   & \textbf{97.93\%}\\ \cline{2-4}
                              & \multirow{3}{*}{Unselected data} & Acc  &  77.61\%\\
                              &                          & S-ratio   &  90.12\%\\ 
                              &                          & S-acc   &  97.63\%\\ \cline{2-4}
                              & \multirow{3}{*}{None} &  Acc&  70.68\%\\
                              &                          & S-ratio   &  78.65\%\\ 
                              &                          & S-acc  &  96.22\%\\ \hline
\end{tabular}
}
}
\end{table}

\noindent\textbf{The influence on the scope of the consistency regularization term.\quad}
As described in Section \ref{section:method}, our co-mix regularization is designed to avoid sample waste. As such, a natural idea is to apply the regularization term to the unselected samples, as in traditional semi-supervised learning. However, our setting differs from traditional semi-supervised learning in that our unselected data only constitutes a small portion of all the data. So we conducted experiments by trying three cases: no regularization term, using the regularization term only for the unselected dataset, and using the regularization term for all examples.

The results in Table \ref{Ablation scope} suggest that with the expansion of the application scope of co-mix regularization term, the model's performance steadily improves, and the number of selected labeled examples also gradually increases. This also shows that our co-mix regularization term and selection strategy can achieve a mutually beneficial effect.

\begin{table}[!t]
\centering
\caption{Accuracy for ablation study on selection criteria.}
\vspace{-0.3cm}
\label{Ablation select criteria}
\resizebox{0.475\textwidth}{!}{
\setlength{\tabcolsep}{3mm}{
\begin{tabular}{c|c|c|c|c}
\toprule
Setting & $t$ & Accuracy & $\gamma$ & Accuracy\\
\midrule
\multicolumn{1}{c|}{\multirow{3}{*}{\makecell[c]{CIFAR-10 \\ $q=0.3$}}}    & $t=2$ &97.03\%& $\gamma=0.80$& 96.24\%  \\
\multicolumn{1}{c|}{} & $t=3$ & 97.50\%& $\gamma=0.90$ & 97.50\% \\
\multicolumn{1}{c|}{} & $t=4$ & 96.15\% & $\gamma=0.95$ & 97.38\%\\
\hline
\multicolumn{1}{c|}{\multirow{3}{*}{\makecell[c]{CIFAR-100 \\ $q=0.1$}}}    & $t=2$ & 82.74\%& $\gamma=0.80$& 80.20\% \\
\multicolumn{1}{c|}{} & $t=3$ & 84.07\%& $\gamma=0.90$ & 84.07\% \\
\multicolumn{1}{c|}{} &$ t=4$ & 83.56\%& $\gamma=0.95$ & 83.56\% \\
\bottomrule
\end{tabular}
}
}
\end{table}

\noindent\textbf{Impact of other parameters on selection criteria.\quad}
Except for the number of selection rules, the two parameters $t$ and $\gamma$ in Eq. (2) and Eq. (3) determine the strictness of our selection criteria. $t$ represents the length of historical prediction stored in $\mathrm{MB}$, while $\gamma$ represents the select threshold for the average prediction confidence of the model for the example prediction in the past $t$ epochs. A larger $t$ and a higher $\gamma$ represent a stricter selection criterion, resulting in a smaller $\mathcal{D}_{\mathrm{sel}}$ size but higher precision, which can affect model training. During the early stages, it may be difficult to select enough examples under the condition of using stricter selection criteria. Therefore, in this experiment, we set the label flipping probability $q$ to 0.3 on CIFAR-10. However, in the later stages, the impact of selection criteria on the final accuracy rate is not significant if the initial stage is passed smoothly. Our experiment shows that $t=3$ and $\gamma=0.9$ are suitable values that can be applied to most experimental environments. The experimental results are presented in Table \ref{Ablation select criteria} and Supplementary Materials.

\begin{figure*}
\centering
  \begin{subfigure}{0.33\linewidth}
    \includegraphics[width=2.2in,height=1.9in]{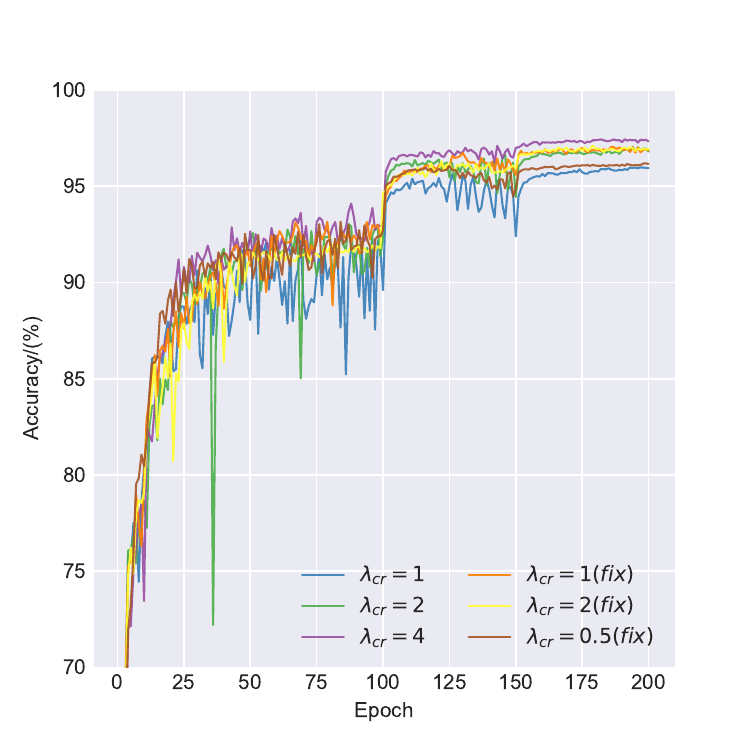}
    \caption{CIFAR-10 Accuracy}
    \label{fig:suba}
  \end{subfigure}
  \begin{subfigure}{0.33\linewidth}
    \includegraphics[width=2.2in,height=1.9in]{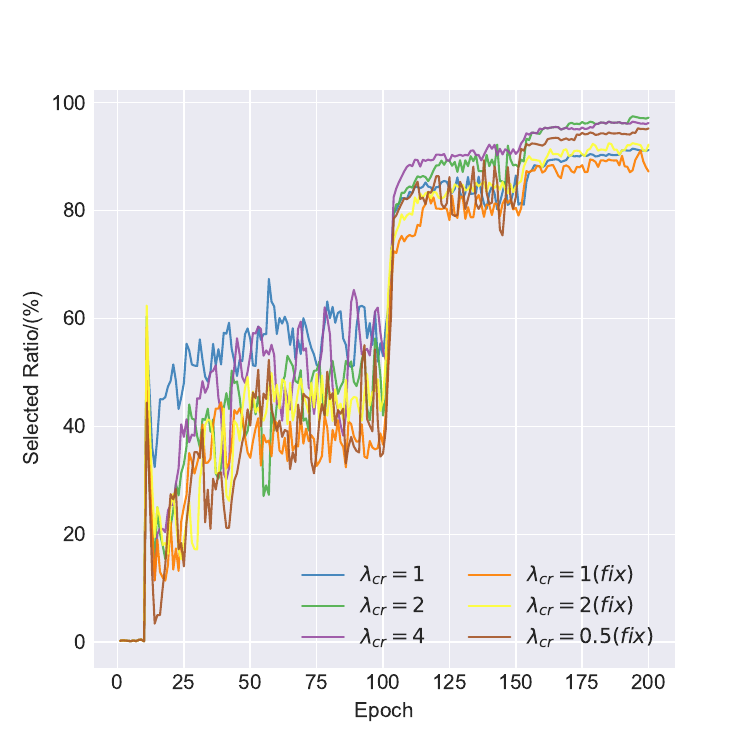}
    \caption{CIFAR-10 Selection Ratio}
    \label{fig:subb}
  \end{subfigure}
\begin{subfigure}{0.33\linewidth}
    \includegraphics[width=2.2in,height=1.9in]{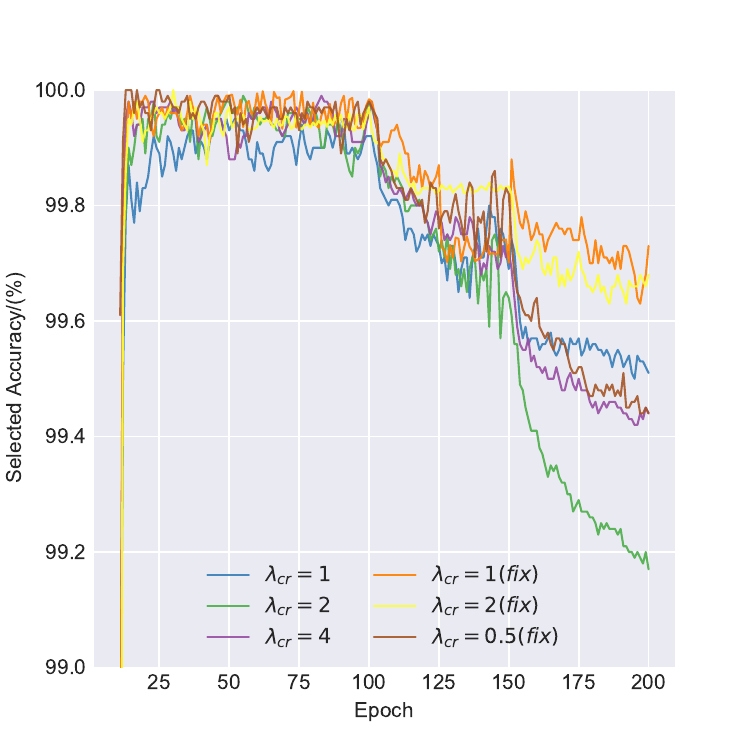}
    \caption{CIFAR-10 Selection Accuracy}
    \label{fig:subc}
  \end{subfigure}\\
\vspace{-0.1cm}
    \begin{subfigure}{0.33\linewidth}
    \includegraphics[width=2.2in,height=1.9in]{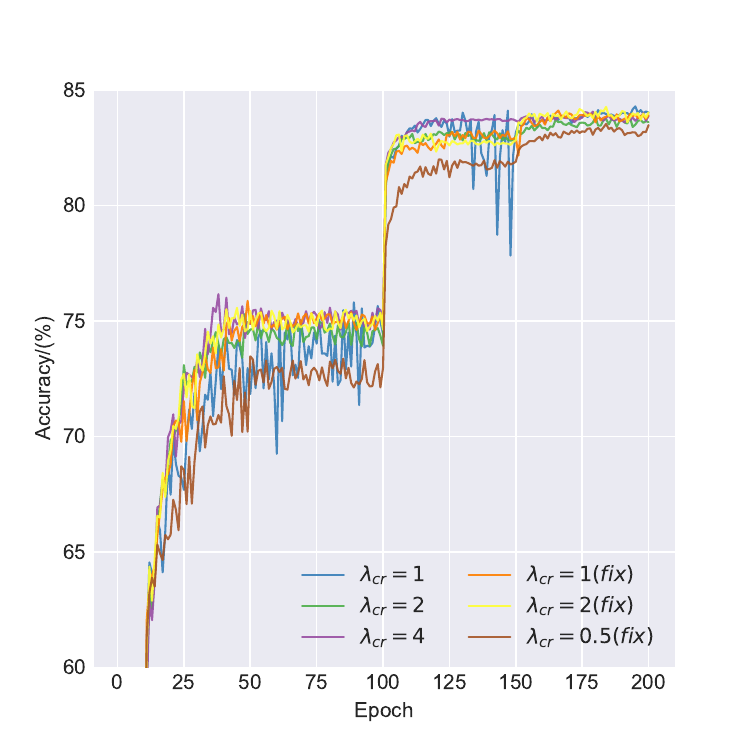}
    \caption{CIFAR-100 Accuracy}
    \label{fig:subd}
  \end{subfigure}
  \begin{subfigure}{0.33\linewidth}
    \includegraphics[width=2.2in,height=1.9in]{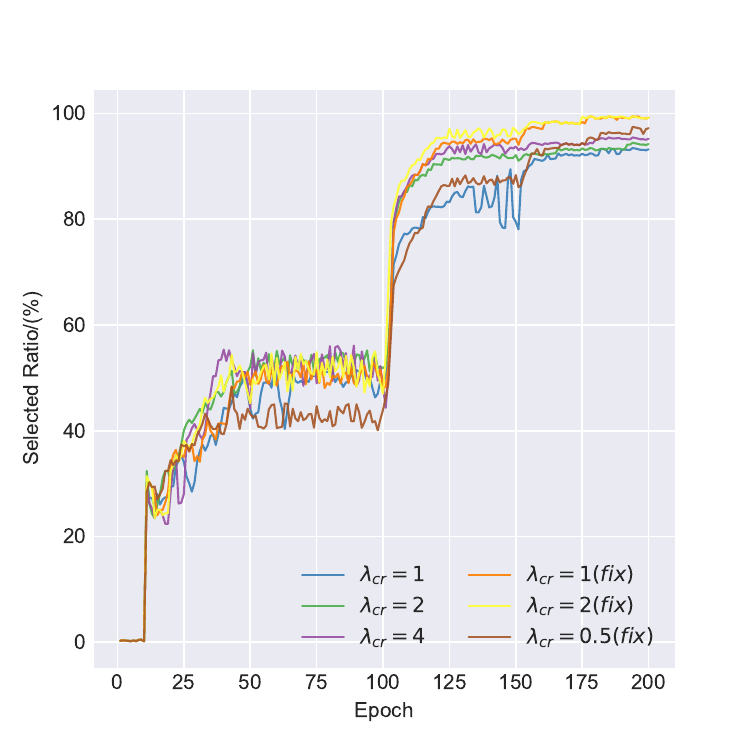}
    \caption{CIFAR-100 Selection Ratio}
    \label{fig:sube}
  \end{subfigure}
\begin{subfigure}{0.33\linewidth}
    \includegraphics[width=2.2in,height=1.9in]{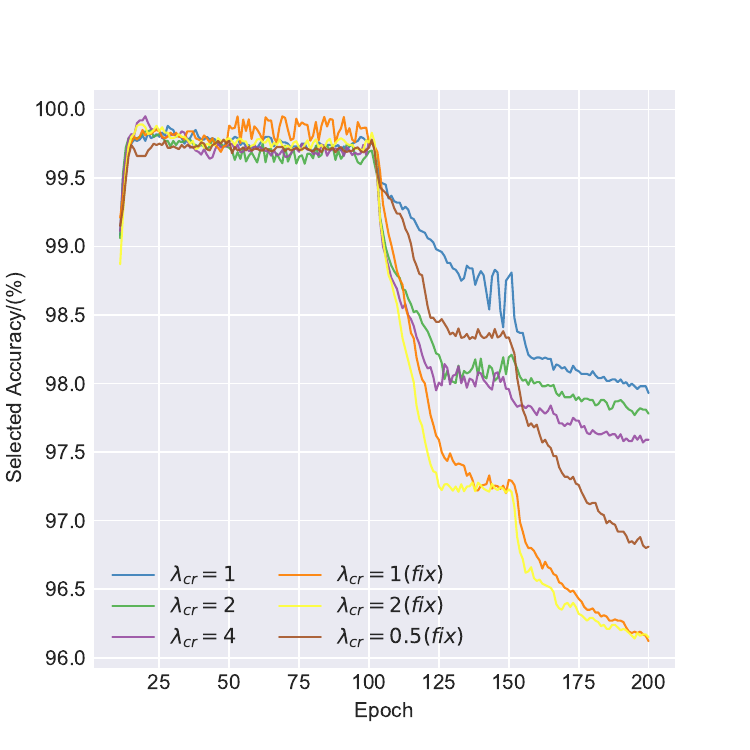}
    \caption{CIFAR-100 Selection Accuracy}
    \label{fig:subf}
  \end{subfigure}
  \caption{Parameter analysis of $\lambda_{\mathrm{cr}}$ on CIFAR-10 and CIFAR-100.}
  \label{fig:mainsub}
\vspace{-0.15cm}
\end{figure*}

\noindent\textbf{Parameter test on $\lambda_{\mathrm{cr}}$.\quad}In this experiment, we test the parameter $\lambda_{\mathrm{d}}$, which weights the contribution of the consistency regularization term to the training loss. As mentioned in  Eq. (14), the parameter $\lambda_{\mathrm{d}}$ is directly influenced by the hyperparameter $\lambda_{\mathrm{cr}}$. Therefore, we test $\lambda_{\mathrm{cr}}=\{1,2,4\}$ on CIFAR-10 ($q=0.5$) and CIFAR-100 ($q=0.1$). At the same time, we also try to fix the parameter $\lambda_{\mathrm{d}}$, that is, its value is not related to the selection ratio $r_{\mathrm{s}}$. This setting we denote by $\lambda_{\mathrm{cr}}\mathrm{(fix)}$, we test $\lambda_{\mathrm{cr}}\mathrm{(fix)}=\{0.5,1,2\}$. The results are visualized in Figure \ref{fig:mainsub}, and detailed data can be found in Supplementary Material.

As mentioned in Table \ref{main scc+sration table} above, CroSel achieves a very high selection ratio. When using a dynamically changing $\lambda_{\mathrm{d}}$, the contribution made by the regularization term would be quite small in the later stages as the learning rate decays. In contrast, with a fixed value of $\lambda_{\mathrm{d}}$, the regularization term would still have a significant contribution in the later stages. Although the accuracy rate on the test set does not show a significant difference between different parameters, the selection effect is critical. A too large contribution of the regularization item will slightly improve the selection ratio at the cost of a decrease in the selection accuracy, which goes against the original intention of our algorithm design. Therefore, we finally decided to use dynamically changing parameters. In other words, we want the algorithm to return to a supervised learning setting with minimal noise as much as possible at the end of the training process.

\begin{figure}
  \begin{subfigure}{0.45\linewidth}
    \includegraphics[width=1.5in,height=1.6in]{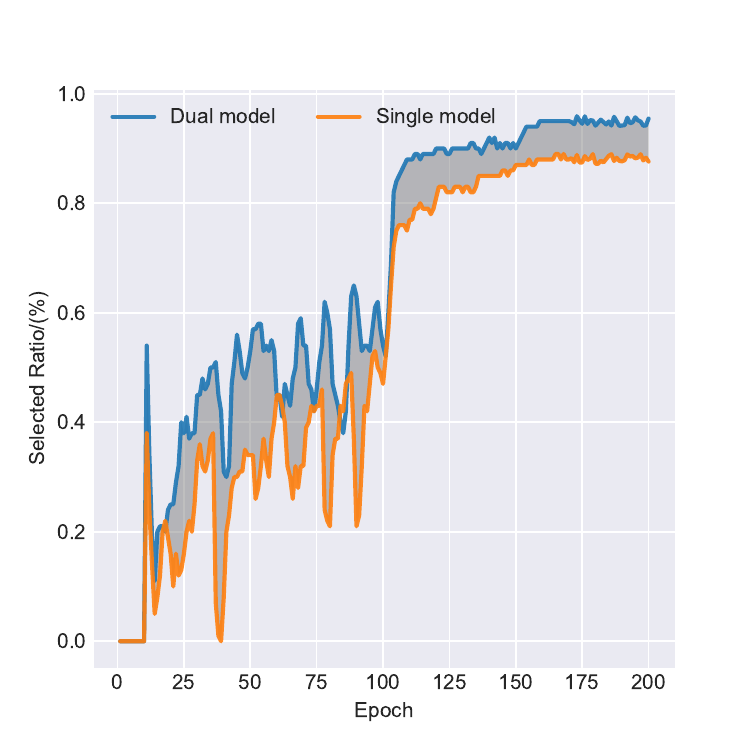}
    \caption{CIFAR-10 comparison.}
    \label{fig:compare-a}
  \end{subfigure}
  \hfill
  \begin{subfigure}{0.45\linewidth}
    \includegraphics[width=1.5in,height=1.6in]{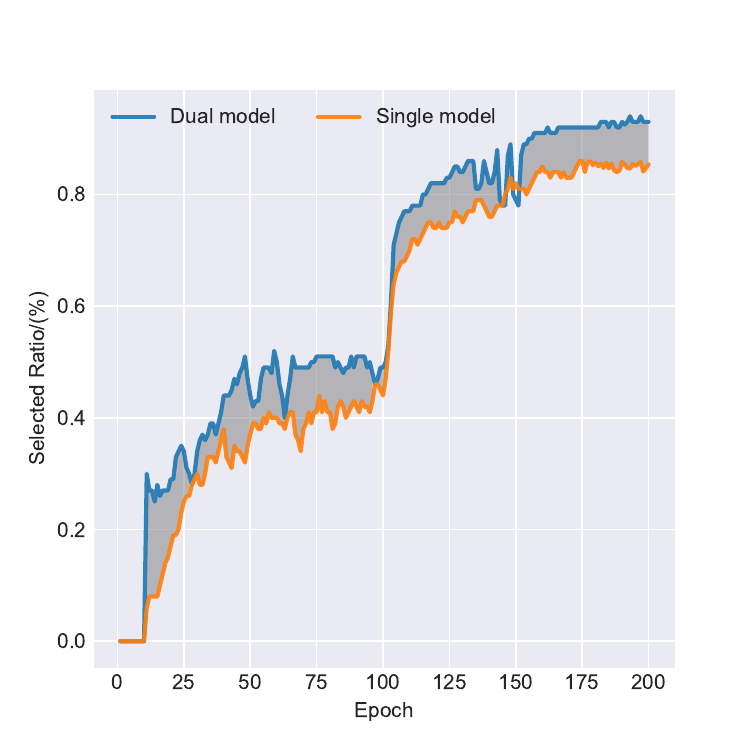}
    \caption{CIFAR-100 comparison.}
    \label{fig:compare-b}
  \end{subfigure}
  \vspace{-0.1cm}
  \caption{Selection ratio comparison between dual model and single model on CIAFR-10 and CIAFR-100.}
  \label{fig:dual}
\vspace{-0.25cm}
\end{figure}

\noindent\textbf{The influence on double model.\quad}As recognized by the community, dual models tend to achieve better performance than single models. We are curious about how effective our selection criteria would be without the adaptive error correction capability of cross selection.  Figure \ref{fig:dual} visualizes the gap in the selection ratio between dual-model and single-model training. It shows that our cross selection strategy can select samples more comprehensively, and on average, about 10\% more training examples can be selected on each dataset. However, the effectiveness of our algorithm is not solely due to the dual model. Even when using a single model with our selection criteria, the accuracy rate on the test set only decreases by 0.83\% and 2.68\% on CIFAR-10 and CIFAR-100 respectively. Furthermore, the high precision of selection is reflected in both settings. Detailed results can be found in Supplementary Material.



\section{Conclusion}
This work introduces CroSel, a novel partial-label learning method that leverages historical prediction to select confident pseudo labels from candidate label sets. The proposed method consists of two parts: a cross selection strategy that enables two models to select ``true" labels for each other, and a consistency regularization term co-mix that avoids sample waste and tiny noise caused by false selection. Empirically, extensive experiments demonstrate the superiority of CroSel, which consistently outperforms previous state-of-the-art methods on multiple benchmark datasets.
\section*{Acknowledgements}
Lei Feng is supported by National Natural Science Foundation of China (Grant No. 62106028) and Chongqing Overseas Chinese Entrepreneurship and Innovation Support Program. Hongxin Wei is supported by Shenzhen Fundamental Research Program JCYJ20230807091809020 and the Center for Computational Science and Engineering at the Southern University of Science and Technology. Yiqun Wang is supported by the Open Projects Program of State Key Laboratory of Multimodal Artificial Intelligence Systems and the NSFC No. 62202076.

{
    \small
    \bibliographystyle{ieeenat_fullname}
    \bibliography{main}
}

\clearpage
\setcounter{page}{1}
\appendix
\maketitlesupplementary

\section{Discussion about the efficiency of CroSel.}
Our method can be viewed as an \textit{example-level} method without interactions between the examples. As a result, it does not suffer from a sharp drop in efficiency as $n$ and $k$ increase. Regarding the update operation, we only need to retain the output of the last layer of the model for each example and update the previous records in the memory bank (MB). This process has a time complexity of $\mathcal{O}(tnk)$. For the selection operation, we just find the maximum and calculate the mean of historical prediction stored in MB of each example, and the time complexity remains $\mathcal{O}(tnk)$.

\section{Experimental details}

\subsection{Datasets}
\begin{itemize}
    \item \textbf{CIFAR-10:} It contains 60,000 $32\times 32$ RGB color pictures in a total of 10 categories. Including 50,000 for the training set and 10,000 for the test set.
    \item \textbf{CIFAR-100:} It has 100 classes, each containing 600 $32\times 32$ RGB color images. Each category has 500 training images and 100 test images. The 100 classes in CIFAR-100 are divided into 20 superclasses. Each image has a ``fine" tag (the class to which it belongs) and a ``rough" tag (the superclass to which it belongs).
    \item \textbf{SVHN:} It is derived from Google Street View door numbers, each image contains a set of Arabic numbers `0-9'. The training set contained 73,257 numbers, the test set 26,032 numbers, and 531,131 additional numbers. Each number is a $32\times 32$ color picture.
\end{itemize}

\subsection{Data augmentations}
Data augmentation is widely used in weakly supervised learning algorithms. There are two types of data augmentations used in our algorithm: ``weak" and ``strong". For ``weak" augmentation, it is just a standard flip-and-shift augmentation strategy consisting of Randomcrop and RandomHorizontalFlip. For ``strong" augmentations, we use the RandAugment strategy for all, which randomly selects the type and magnitude of data augmentation with the same probability.


\subsection{Compared methods}
We reimplement CC, PRODEN, LWS, and CRDPLL using the same training scheme as CroSel. For PiCO and POP, we just follow their original training schemes and change the backbone to WRN-34-10. For a fair comparison, we add weak augmentation to those methods that are not equipped with data augmentation in their original scheme (CC, PRODEN, LWS). For other methods already equipped with augmentation (POP, PiCO, CRDPLL), we follow their settings illustrated in their paper.

\subsection{Implementations}
We set the batch size as 64 and total epochs as 200, using SGD as optimizer with a momentum of 0.9, and set the initial learning rate as 0.1, which is divided by 10 after 100 and 150 epochs respectively. For the hyper-parameters in our method, we set $t=3$, $\alpha=0.75$, $T=0.5$ for all datasets, and $\lambda_{\mathrm{cr}}=1$ for CIFAR-100, $\lambda_{\mathrm{cr}}=4$ for others. For the selection threshold, we set $\gamma=0.9$ for CIFAR-type datasets, and $\gamma=0.85$ for SVHN. 

\subsection{Detailed results and more ablation experiment}
Due to constraints on page space, we are unable to display all experimental results within the main text. Therefore, we focus on presenting the most crucial metrics or visualization outcomes. Specific experimental results and some more ablation experiments will be showcased in this section.

\noindent\textbf{Discussion about the influence of parameter $t$ and $\gamma$.} 
As we mentioned before, $t$ represents the length of historical prediction stored in $\mathrm{MB}$, while $\gamma$ represents the select threshold for the average prediction confidence of the model for the example prediction in the past $t$ epochs.
The two parameters determine the strictness of our selection criteria. 

In the main text, we exclusively presented the final test accuracy. However, here we provide additional insights by showcasing the number of selection and selection accuracy in Tables \ref{Detailed results for Parameter test on $t$} and \ref{Detailed results for Parameter test on gamma}. Regarding the parameter $t$, an increase in the storage length of the memory bank generally signifies a more stringent selection criterion. 
This results in a higher loss of selection numbers in exchange for a slight improvement in selecting accuracy. For the parameter $\gamma$, the model is more sensitive to some changes on this threshold. Lowering the threshold tends to introduce more noise, resulting in a decline in model performance, which subsequently affects the selection accuracy.

\begin{table}[h]
\caption{Detailed results for Parameter test on $t$.}
\vspace{-0.5cm}
\label{Detailed results for Parameter test on $t$}
\begin{center}
\begin{tabular}{l|c|ccc}
\toprule
Setting & $t$ & Accuracy & S-ratio &S-acc\\
\midrule
\multicolumn{1}{c|}{\multirow{3}{*}{\makecell[c]{CIFAR-10 \\ $q=0.3$}}}    &$t=2$& 97.03\% & 99.07\% &99.62\% \\
\multicolumn{1}{c|}{} & $t=3$ & 97.50\% & 98.10\% & 99.55\% \\
\multicolumn{1}{c|}{} & $t=4$ & 96.15\% & 89.23\%& 99.76\% \\
\hline
\multicolumn{1}{c|}{\multirow{3}{*}{\makecell[c]{CIFAR-100 \\ $q=0.1$}}}    & $t=2$ &82.74\% & 87.32\% &98.50\% \\
\multicolumn{1}{c|}{} & $t=3$ & 84.07\% & 93.61\% & 97.93\% \\
\multicolumn{1}{c|}{} & $t=4$ & 83.56\% & 93.24\% &98.23\% \\
\bottomrule
\end{tabular}
\end{center}
\end{table}

\begin{table}[h]
\caption{Detailed results for Parameter test on 
$\gamma$.}
\vspace{-0.5cm}
\label{Detailed results for Parameter test on gamma}
\begin{center}
\scalebox{0.95}{
\begin{tabular}{l|c|ccc}
\toprule
Setting & $\gamma$ & Accuracy & S-ratio &S-acc\\
\midrule
\multicolumn{1}{c|}{\multirow{3}{*}{\makecell[c]{CIFAR-10 \\ $q=0.3$}}}    &$\gamma=0.8$& 96.24\% & 95.29\% &99.08\% \\
\multicolumn{1}{c|}{} & $\gamma=0.9$ & 97.50\% & 98.10\% & 99.55\% \\
\multicolumn{1}{c|}{} & $\gamma=0.95$ & 97.38\% & 93.10\%& 99.87\% \\
\hline
\multicolumn{1}{c|}{\multirow{3}{*}{\makecell[c]{CIFAR-100 \\ $q=0.1$}}}    & $\gamma=0.8$ &80.20\% & 82.11\% &98.63\% \\
\multicolumn{1}{c|}{} & $\gamma=0.9$ & 84.07\% & 93.61\% & 97.93\% \\
\multicolumn{1}{c|}{} & $\gamma=0.95$ & 81.23\% & 67.32\% & 99.26\% \\
\bottomrule
\end{tabular}
}
\end{center}
\end{table}

\noindent\textbf{Detailed results for comparison between Dual model and Single model.} 
In the main text, we only displayed the visual graph illustrating the number of selection. Here, we provide specific values for multiple metrics in Table \ref{Detailed results of comparison between dual model and single model}. On one hand, the dual model substantially increases the number of selected pseudo-labeled samples without diminishing the selection accuracy, which demonstrates the effectiveness of the dual model. On the other hand, even without the utilization of the dual model, we do not observe a sharp decrease in test accuracy, highlighting the stability of CroSel, whose performance improvement does not rely solely on the dual model.

\begin{table}[h]
\caption{Detailed results of comparison between Dual model and Single model.}
\label{Detailed results of comparison between dual model and single model}
\vspace{-0.5cm}
\begin{center}
\scalebox{0.9}{
\begin{tabular}{c|c|ccc}
\toprule
Setting & Model & Accuracy & S-ratio & S-acc\\
\midrule
\multicolumn{1}{c|}{\multirow{2}{*}{\makecell[c]{CIFAR-10 \\ $q=0.5$}}}    &Single model& 96.51\% &87.61\% & 99.72\%  \\
\multicolumn{1}{c|}{} & Dual model& 97.34\% &96.25\%& 99.44\%  \\
\hline
\multicolumn{1}{c|}{\multirow{2}{*}{\makecell[c]{CIFAR-100 \\ $q=0.1$}}}    & Single model&81.39\% &85.39\% & 98.35\%  \\
\multicolumn{1}{c|}{} & Dual model& 84.07\%& 93.61\%& 97.93\%  \\
\bottomrule
\end{tabular}
}
\end{center}
\end{table}

\noindent\textbf{Detailed results for Parameter test on $\lambda_{\mathrm{cr}}$.} 
As mentioned earlier, $\lambda_{\mathrm{d}}$ weights the contribution of the consistency regularization term to the training loss. As described in Eq. (14), the parameter $\lambda_{\mathrm{d}}$ is directly influenced by the hyperparameter $\lambda_{\mathrm{cr}}$. We tested $\lambda_{\mathrm{cr}}$ values of \{1, 2, 4\} and $\lambda_{\mathrm{cr}}\mathrm{(fix)}$ values of \{0.5, 1, 2\}.

In the main text, we visualized the evolution of various metrics as the training epoch progresses. The specific values at the end of training are provided in this section. Table \ref{Detailed:Ablation lambda-cr results} serves as an effective demonstration of our algorithm's robustness to the parameter $\lambda_{\mathrm{cr}}$.
Using dynamically varying $\lambda_{\mathrm{cr}}$ tends to result in higher selection accuracy compared to using a fixed value of $\lambda_{\mathrm{cr}}$, thereby impacting the model performance. This observation aligns with our original intention of reducing the contribution of regularization terms in the final loss towards the end of training, thereby transitioning the model back to a simpler supervised learning scenario.

\begin{table}[h]
\caption{Detailed results for Parameter test on $\lambda_{\mathrm{cr}}.$}
\vspace{-0.5cm}
\label{Detailed:Ablation lambda-cr results}
\begin{center}
\scalebox{0.87}{
\begin{tabular}{l|c|ccc}
\toprule
Setting & $\lambda_{cr} $& Accuracy & S-ratio &S-acc\\
\midrule
\multicolumn{1}{c|}{\multirow{6}{*}{\makecell[c]{CIFAR-10 \\ $q=0.5$}}}    &$\lambda_{\mathrm{cr}}=1$& 95.94\% & 91.57\% &99.51\% \\
\multicolumn{1}{c|}{} & $\lambda_{\mathrm{cr}}=2$ & 96.80\% & 97.32\% & 99.17\% \\
\multicolumn{1}{c|}{} & $\lambda_{\mathrm{cr}}=4$ & 97.33\% & 96.25\% & 99.44\% \\
\multicolumn{1}{c|}{} & $\lambda_{\mathrm{cr}}=1 (\mathrm{fixed})$ & 96.88\% & 87.88\% & 99.73\% \\
\multicolumn{1}{c|}{} & $\lambda_{\mathrm{cr}}=2 (\mathrm{fixed})$ & 96.95\% & 92.19\% & 99.68\% \\
\multicolumn{1}{c|}{} & $\lambda_{\mathrm{cr}}=0.5 (\mathrm{fixed})$ & 96.16\% & 95.21\% & 99.44\% \\
\hline
\multicolumn{1}{c|}{\multirow{6}{*}{\makecell[c]{CIFAR-100 \\ $q=0.1$}}}    &$\lambda_{\mathrm{cr}}=1$& 84.07\% & 93.61\% &97.93\% \\
\multicolumn{1}{c|}{} & $\lambda_{\mathrm{cr}}=2$ & 83.61\% & 94.15\% & 97.78\% \\
\multicolumn{1}{c|}{} & $\lambda_{\mathrm{cr}}=4$ & 83.88\% & 95.63\% & 97.59\% \\
\multicolumn{1}{c|}{} & $\lambda_{\mathrm{cr}}=1 (\mathrm{fixed})$ & 83.91\% & 99.35\% & 96.12\% \\
\multicolumn{1}{c|}{} & $\lambda_{\mathrm{cr}}=2 (\mathrm{fixed})$ & 84.03\% & 99.17\% & 96.15\% \\
\multicolumn{1}{c|}{} & $\lambda_{\mathrm{cr}}=0.5 (\mathrm{fixed})$ & 83.48\% & 97.02\% & 96.81\% \\
\bottomrule
\end{tabular}
}
\end{center}
\end{table}


\noindent\textbf{Influence for the data augmentation on $D_{\mathrm{sel}}$.} 
Data augmentation plays a crucial role in weakly supervised learning. However, our selection criteria rely on historical prediction to select examples with high confidence.
Consequently, we cannot guarantee that employing stronger data augmentation strategy will invariably yield superior results and selection effects.
The randomness and variability inherent in data augmentation may introduce some adverse effects on the model's memorization capabilities, particularly as the strength of the augmentation strategy increases. Therefore, in this section, we delve into the impact of data augmentation on $\mathcal{D}_{\mathrm{sel}}$ and its influence on label selection and overall training dynamics.

As shown in Table \ref{Detailed results for Parameter test on dsel}, weak augmentation is a more appropriate and effective choice. However, because of the impact of consistency regularization items, even if data augmentation is not used on $\mathcal{D}_{\mathrm{sel}}$, there is no significant performance degradation. It is noting that strong data augmentation has an adverse effect on the selection of examples, especially in CIFAR-100, which may be related to the fact that the historical predictions stored in $\mathrm{MB}$ are produced by data that have not been augmented.

\begin{table}[!htpb]
\caption{Detailed results for Parameter test on $\mathcal{D}_{\mathrm{sel}}.$}
\vspace{-0.5cm}
\label{Detailed results for Parameter test on dsel}
\begin{center}
\scalebox{0.85}{
\begin{tabular}{l|c|ccc}
\toprule
Setting & Data augmentation& Accuracy & S-ratio &S-acc\\
\midrule
\multicolumn{1}{c|}{\multirow{3}{*}{\makecell[c]{CIFAR-10 \\ $q=0.3$}}}    &None& 96.91\% & 97.44\% &99.60\% \\
\multicolumn{1}{c|}{} & Weak & 97.50\% & 98.10\% & 99.55\% \\
\multicolumn{1}{c|}{} & Strong& 97.22\% & 96.26\%& 99.75\% \\
\hline
\multicolumn{1}{c|}{\multirow{3}{*}{\makecell[c]{CIFAR-100 \\ $q=0.1$}}}    & None &83.74\% & 90.48\% &98.32\% \\
\multicolumn{1}{c|}{} & Weak& 84.07\% & 93.61\% & 97.93\% \\
\multicolumn{1}{c|}{} & Strong & 81.01\% & 79.16\% & 98.98\% \\
\bottomrule
\end{tabular}
}
\end{center}
\end{table}

\end{document}